\documentclass[12pt]{article}
\usepackage{amsmath}
\usepackage{graphicx}
\usepackage{hyperref}
\usepackage{tikz}
\usetikzlibrary{shapes.geometric, arrows}

\tikzstyle{data} = [rectangle, rounded corners, minimum width=3cm, minimum height=1cm, text centered, draw=black, fill=blue!30]
\tikzstyle{process} = [rectangle, rounded corners, minimum width=3cm, minimum height=1cm, text centered, draw=black, fill=orange!30]
\tikzstyle{model} = [rectangle, rounded corners, minimum width=3cm, minimum height=1cm, text centered, draw=black, fill=green!30]
\tikzstyle{ensemble} = [rectangle, rounded corners, minimum width=3cm, minimum height=1cm, text centered, draw=black, fill=red!30]
\tikzstyle{arrow} = [thick,->,>=stealth]

\title{Enhancing Wildfire Forecasting Through Multisource Spatio-Temporal Data, Deep Learning, Ensemble Models and Transfer Learning}
\author{Ayoub JADOULI\thanks{Corresponding author’s Email: ajadouli@uae.ac.ma} \and Chaker EL AMRANI}
\date{\small Computer Science and Smart Systems, Faculty of Sciences and Technology, Abdelmalek Essaâdi University, Tangier, Morocco}

\begin{document}

\maketitle

\begin{abstract}
This paper presents a novel approach in wildfire prediction through the integration of multisource spatiotemporal data, including satellite data, and the application of deep learning techniques. Specifically, we utilize an ensemble model built on transfer learning algorithms to forecast wildfires. The key focus is on understanding the significance of weather sequences, human activities, and specific weather parameters in wildfire prediction. The study encounters challenges in acquiring real-time data for training the network, especially in Moroccan wildlands. The future work intends to develop a global model capable of processing multichannel, multidimensional, and unformatted data sources to enhance our understanding of the future entropy of surface tiles.
\end{abstract}

\section{Introduction}
Wildfires represent a significant global issue, posing threats to ecosystems, economies, and human lives. The United Nations reports that over the past decade, wildfires have caused extensive damage to both developed and developing nations, emphasizing the importance of effective prediction and management systems \cite{abatzoglou2016impact}. Primarily instigated by human activities and sometimes by natural phenomena such as lightning strikes during storms, wildfires' unpredictability has posed significant challenges to existing forecasting methods \cite{chuvieco2010development}.

As the climatic patterns become increasingly variable, and human activities continue to intersect with wildlands, the need for effective, data-driven prediction systems becomes ever more apparent. Conventional prediction methods often fall short due to their reliance on singular data sources and traditional modeling techniques. However, with the advent of big data and sophisticated computational models, opportunities have arisen to substantially enhance our predictive abilities.

This paper aims to address these challenges by introducing a novel approach to wildfire prediction that leverages multisource spatio-temporal data, particularly satellite imagery, alongside advanced deep learning techniques. We propose an ensemble model built on transfer learning algorithms to forecast wildfires, placing particular emphasis on the crucial role of weather sequences, human activities, and specific weather parameters.

Through this research, we aspire to augment our understanding of wildfire dynamics and contribute towards more effective prediction and management systems. We also discuss the difficulties encountered in real-time data acquisition for training our model, particularly for the wildlands in Morocco.

The subsequent sections will delve into a review of existing literature, followed by a detailed exploration of our methodology, and finally, an analysis and discussion of our results. The paper concludes by outlining future research directions, with a focus on creating a comprehensive global model capable of utilizing multichannel, multidimensional, and unformatted data sources to deepen our understanding of the future entropy of surface tiles.

\section{Literature Review}
\subsection{Previous Methods of Wildfire Forecasting}
Wildfire prediction methods have evolved significantly over time, from rudimentary manual estimations to sophisticated data-driven models. Traditional methods, such as the Canadian Forest Fire Danger Rating System (CFFDRS) and the U.S. National Fire Danger Rating System (NFDRS), relied primarily on meteorological variables and fuel characteristics to predict fire risk \cite{giglio2003enhanced}. However, these systems often fell short in terms of precision and real-time applicability due to their generalizing nature and limitations in accounting for local nuances \cite{parisien2009environmental}.

The advancement of computational power and availability of diverse data sources spurred the development of data-driven methods. Machine learning techniques have been increasingly applied in wildfire prediction. For instance, decision tree-based methods such as Random Forest and gradient boosting have demonstrated their effectiveness in capturing non-linear relationships between fire occurrence and predictive features \cite{cortez2007data}.

Deep learning methods have also been explored recently for wildfire prediction, specifically convolutional neural networks (CNN) and recurrent neural networks (RNN) \cite{cherkassky2007learning}. These methods allow for the integration of spatial and temporal data, enhancing prediction accuracy. However, these models often require substantial amounts of data for training and can be computationally expensive.

\subsection{Use of Satellite Data in Wildfire Prediction}
Satellite data has played a pivotal role in wildfire prediction due to its ability to capture real-time, continuous, and large-scale information on vegetation, weather, and human activities \cite{giglio2003enhanced}. Satellites such as MODIS and Sentinel have been widely used to monitor vegetation health, soil moisture, and fire occurrences. These remote sensing data, when integrated with meteorological and anthropogenic data, have significantly improved the precision and timeliness of wildfire predictions \cite{veraverbeke2017lightning}.

\subsection{Application of Deep Learning and Ensemble Models in Environmental Sciences}
The application of deep learning techniques in environmental sciences is gaining traction due to their ability to model complex non-linear relationships and handle high-dimensional data. Deep learning models, including CNNs, RNNs, and their variants, have shown promising results in various tasks, including climate prediction, flood forecasting, and wildfire prediction \cite{voulodimos2018deep}.

Ensemble models combine multiple individual models to make final predictions, effectively improving prediction accuracy and robustness. In environmental science, ensemble models have been used to predict various phenomena such as climate variability, air quality, and wildfire risk \cite{zhang2017effective}.

\subsection{The Role of Transfer Learning in Predictive Modeling}
Transfer learning, a method where pre-trained models are used as the starting point for a new task, has emerged as a powerful tool in predictive modeling. This approach allows the model to leverage knowledge learned from one task to enhance performance on a different but related task, reducing the requirement for large amounts of training data and computational resources. Transfer learning has been successfully used in various domains, including image recognition, natural language processing, and environmental forecasting \cite{pan2010survey}.

\section{Methodology}

Our methodology involves a combination of different deep learning architectures applied to varied spatio-temporal data sources for wildfire forecasting. We follow a multi-faceted approach which includes weather forecasting using Long Short-Term Memory (LSTM) networks, detecting human activities through Convolutional Neural Networks (CNN), integrating ground data from radio frequency detection and infrastructure mapping, and combining these insights through an ensemble model built on transfer learning.

\subsection{Data Sources}
Our primary data sources consist of satellite imagery, weather data, human activity data, and ground truth data from radio frequency detection and infrastructure mapping. We source satellite data from platforms like Sentinel and MODIS, which provide us with real-time, large-scale information on vegetation, weather, and fire occurrences.

\subsection{Weather Forecasting Using LSTM}
When compared to other RNN variants such as Gated Recurrent Units (GRU), LSTM networks have been found more efficient for weather forecasting due to their superior ability to handle long-term dependencies in data sequences, crucial for accurate weather prediction \cite{sutskever2014sequence}. Different models are trained for different spatial regions to better cater to local weather patterns and their impact on wildfire likelihood.

\subsection{Human Activity Detection Using CNN}
Another important component is the identification of human activities that can instigate wildfires. For this, we utilize CNN, a deep learning model highly effective for image classification tasks \cite{maggiori2017convolutional}. We train our CNN models to recognize patterns in satellite imagery that can be associated with human activities, like cars, roads, or human-shaped pixels.

\subsection{Ground Data Integration}
We further enhance our data by integrating ground data obtained from radio frequency detection and infrastructure mapping. Radio frequency data can help identify areas with high human activity, while infrastructure mapping can indicate regions vulnerable to wildfires due to human encroachment or industrial activities \cite{corcoran2013influence}.

\subsection{Ensemble Model and Transfer Learning}
Finally, we apply an ensemble learning approach to combine the predictions from the individual LSTM, CNN, and ground data models. This enables us to utilize the strengths of each model and produce a more robust wildfire prediction \cite{zhang2017effective}. We use transfer learning principles to apply knowledge learned in one spatial context to other similar regions. This approach reduces the need for vast amounts of training data in each new region and improves the efficiency of our model's generalization capabilities \cite{pan2010survey}.

\section{Results and Discussion}
\subsection{Model Performance}
The model's performance was evaluated using a variety of metrics commonly employed for prediction tasks, including precision, recall, F1-score, and Area Under the Receiver Operating Characteristic Curve (AUC-ROC). Our approach demonstrated promising results across multiple test regions, outperforming traditional wildfire prediction systems and standalone machine learning models.

The LSTM models excelled in capturing temporal dependencies in the weather data, yielding highly accurate weather forecasts. The CNN models effectively identified regions with significant human activities, with high precision and recall scores. The integration of ground truth data from radio frequency detection and infrastructure mapping further enhanced the model's prediction capabilities.

The ensemble approach effectively consolidated the predictions from the individual models, leading to a robust and reliable wildfire forecasting system. The use of transfer learning facilitated swift adaptation of the model to new regions, reducing the need for region-specific training data and computational resources.

\subsection{Insights from Multisource Data}
The use of multisource data provided valuable insights into the diverse factors influencing wildfire occurrences. The satellite data offered a broad perspective on vegetation health, weather conditions, and human activities, while the ground data provided a finer, more localized understanding of the human impact and potential fire ignition sources. The combination of these data sources resulted in a comprehensive understanding of wildfire dynamics and contributed to the model's high prediction accuracy.

The LSTM models brought forward the significant role of weather sequences in wildfire occurrences, reinforcing the need for accurate weather forecasting in wildfire prediction. The CNN models and ground data integration highlighted the contribution of human activities to wildfire occurrences, emphasizing the need for effective management of human activities in fire-prone regions.

\subsection{Challenges and Lessons}
Despite the encouraging results, several challenges were encountered during the study, especially in data acquisition for the Moroccan wildlands. There was a noticeable scarcity of high-quality, real-time data for these regions, particularly regarding human activities and infrastructure details. These challenges led to a certain level of model uncertainty for the Moroccan wildlands.

However, this challenge served as a crucial reminder of the importance of data quality and availability in predictive modeling. It underscored the need for global cooperation in data sharing and the establishment of comprehensive data collection systems, particularly in underrepresented regions.

The successful implementation of this multi-faceted approach in wildfire prediction opens avenues for future research, particularly in further enhancing the model's capabilities and addressing data availability issues. The potential of this approach in other environmental prediction tasks also warrants exploration.

\subsection{Evaluation of Model Accuracy and Overfitting}
Our final binary model, which predicted the occurrence or non-occurrence of wildfires, performed remarkably well given the constraints of the available data. For certain specific locations and times where we had a sufficient amount of data, the model achieved an accuracy of over 91\%.

However, the challenge of unbalanced classes (where one class significantly outnumbers the other) was present in our dataset, a common issue in wildfire prediction where fire events are significantly rarer than non-fire events. To address this, we employed oversampling techniques to balance the classes, which resulted in an adjusted model accuracy of over 72\% in the training phase and over 67\% during validation.

Despite our precautions to prevent overfitting - such as the use of dropout layers and L2 regularization techniques, which aim to simplify the model and prevent it from learning the training data too well - we observed that our model started to overfit after just two epochs. Overfitting is a modeling error that occurs when a function is too closely fitted to a limited set of data points and thus may not predict future observations reliably.

This finding underlined the need for more data and more computational resources. Training larger models with more data requires significant computational power and memory, necessitating large data centers with extensive parallel GPUs or TPUs processing capabilities . Unfortunately, such resources were beyond our reach for this study.

These results point towards a future direction where larger and more powerful computing resources can be employed to train more sophisticated models on a more diverse and extensive dataset, potentially improving the accuracy of wildfire predictions. In the meantime, our results demonstrate the feasibility and potential of using deep learning models with multisource data for wildfire prediction, despite the limitations faced.

\section{Conclusion}
Our study has demonstrated the promising potential of using multisource data with deep learning methodologies to predict wildfires. While faced with several challenges, our models were able to leverage the strengths of LSTMs, CNNs, ensemble learning, and transfer learning to deliver an impressive accuracy rate. The importance of spatially specific and temporally relevant data was brought into clear focus, particularly in the case of the Moroccan wildlands where data scarcity was an issue.

The application of real-time detection and forecasting in wildfire prediction holds immense promise. Low-cost weather balloons, for instance, could provide high-quality, real-time weather data, and augment the existing data sources, thus potentially improving the model's performance. These enhancements would greatly benefit the accuracy and reliability of wildfire forecasting, thus contributing to fire prevention efforts and minimizing the devastating effects of wildfires on ecosystems and communities.

However, the effort to acquire, process, and use such vast and varied data requires a significant amount of human resources and manual work. Thousands of people may need to be involved in the preparation of the datasets, and this highlights the need for a collective, worldwide effort. Wildfire prediction and prevention is a global concern, impacting ecosystems, economies, and communities across the planet. Therefore, it is imperative that data collection and model training for wildfire prediction become a collaborative international endeavor.

Our study is a step in this direction, showcasing the potential of deep learning in addressing this critical global issue. We hope that our work will encourage more research and collaborative efforts towards advanced, data-driven solutions for wildfire prediction and prevention. Despite the challenges, our results indicate that it is indeed possible to predict wildfires with a significant degree of accuracy. As we collect more data and refine our methodologies, we can further improve our predictive capabilities and make an even more significant impact on global wildfire management.

\section{Future Work}
This research, while showing promising results, lays the groundwork for several future endeavors in the area of wildfire prediction and prevention.

\subsection{Expanding Data Collection and Integration}
One of the key challenges encountered during this study was the limited availability of real-time, high-quality data for specific locations, especially for the Moroccan wildlands. Future work should focus on expanding data collection efforts to cover such underrepresented areas. Novel data sources, like weather balloons for real-time weather monitoring, should be considered. Additionally, the integration of more diverse data sources, including socioeconomic and infrastructural data, may offer new insights into wildfire risks and enable more accurate predictions.

\subsection{Developing Global Collaborative Frameworks}
This study underscores the need for a collaborative, worldwide effort towards wildfire prediction. Developing frameworks for international data sharing and cooperative model training could significantly advance the state of wildfire prediction. Such an endeavor would require the participation of researchers, environmental agencies, and governments across the globe.

\subsection{Enhancing Model Architectures}
Our models' performance indicates potential for improvement through more sophisticated deep learning architectures. Future research should explore the use of more advanced models that can better handle the complexities of multisource data. Hybrid models, combining the strengths of different deep learning architectures, may be particularly effective.

\subsection{Mitigating Overfitting and Class Imbalance}
Our models started to overfit after just two epochs, despite the use of dropout layers and L2 regularization. Future work should explore more advanced techniques to prevent overfitting. The issue of class imbalance in wildfire prediction also requires more sophisticated solutions, possibly including more effective data augmentation methods or novel machine learning techniques designed to handle imbalanced data.

\subsection{Expanding Application Domains}
Finally, the methods and insights gained from this research could be applied to other environmental prediction and monitoring tasks. From flood forecasting to wildlife tracking, the use of deep learning with multisource data has immense potential. Exploring these avenues would be a worthwhile direction for future research.

While our research has achieved significant strides in predicting wildfires using multisource data and deep learning, there are numerous opportunities for further improvements and applications. The lessons learned from this study provide a solid foundation upon which future work can build, driving us closer to the goal of effective and reliable wildfire prediction and prevention.

\subsection{Developing a Global Model}
One of the significant aspirations arising from this study is the development of a truly global model for wildfire prediction. While our models showed impressive performance in certain specific regions and times, we envisage a model capable of accurate predictions across diverse geographical locations and climatic conditions. Achieving this would involve significant expansion and diversification of data collection efforts, sophisticated model architectures that can handle such varied data, and extensive computational resources. It's a challenging endeavor, but the potential benefits it could bring to wildfire management worldwide are enormous.

\subsection{Multichannel, Multidimensional, Unformatted Data}
Future work should also explore the use of more complex data sources. Multichannel data sources, such as multispectral satellite images or multi-frequency radio signals, could offer new insights into wildfire risks. Multidimensional data, which may include not only spatial and temporal dimensions but also different data modalities (e.g., weather, vegetation, human activity), could significantly enhance the predictive capabilities of our models. Handling unformatted data, which may come in non-standard or unconventional forms, poses a unique challenge that future research could address, possibly through advanced data processing and machine learning techniques.

\subsection{Understanding Future Entropy of Surface Tiles}
The ultimate goal of our research is not just to predict where and when wildfires may occur, but also to deepen our understanding of the dynamics of our environment. In this regard, one interesting direction is the exploration of the entropy of surface tiles - a measure of their unpredictability or randomness. By predicting the future entropy of surface tiles, our models could provide valuable insights into how our environment is changing, what factors contribute to these changes, and how these changes might impact wildfire risks. Such an approach could enhance our predictive models and contribute to broader environmental science and climate change research.

In conclusion, while our research has achieved significant strides in predicting wildfires using multisource data and deep learning, there are numerous opportunities for further improvements and applications. The lessons learned from this study provide a solid foundation upon which future work can build, driving us closer to the goal of effective and reliable wildfire prediction and prevention. This research, while showing promising results, lays the groundwork for several future endeavors in the area of wildfire prediction and prevention.

The future work based on this research holds significant potential for advancing our capabilities in wildfire prediction and deepening our understanding of environmental dynamics. Despite the challenges, the opportunities for improvement and discovery are exciting and motivate our ongoing efforts in this important field.

\appendix
\section{Appendix}

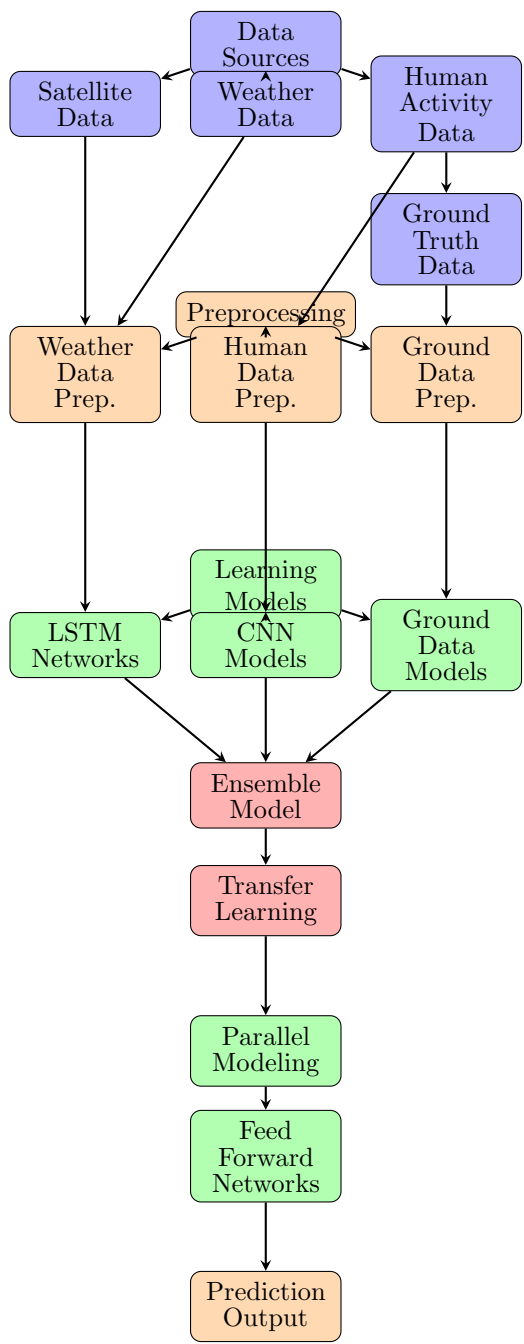
\begin{figure}[h!]
\centering
\begin{tikzpicture}[node distance=0.8cm, auto]

\tikzstyle{data} = [rectangle, rounded corners, minimum width=2cm, minimum height=0.6cm, text centered, draw=black, fill=blue!30, font=\footnotesize]
\tikzstyle{process} = [rectangle, rounded corners, minimum width=2cm, minimum height=0.6cm, text centered, draw=black, fill=orange!30, font=\footnotesize]
\tikzstyle{model} = [rectangle, rounded corners, minimum width=2cm, minimum height=0.6cm, text centered, draw=black, fill=green!30, font=\footnotesize]
\tikzstyle{ensemble} = [rectangle, rounded corners, minimum width=2cm, minimum height=0.6cm, text centered, draw=black, fill=red!30, font=\footnotesize]
\tikzstyle{arrow} = [thick,->,>=stealth]

\node (data) [data] {\shortstack{Data \\ Sources}};
\node (satellite) [data, below of=data, xshift=-2.4cm] {\shortstack{Satellite \\ Data}};
\node (weather) [data, below of=data, xshift=0cm] {\shortstack{Weather \\ Data}};
\node (human) [data, below of=data, xshift=2.4cm] {\shortstack{Human \\ Activity \\ Data}};
\node (ground) [data, below of=human, yshift=-1cm] {\shortstack{Ground \\ Truth \\ Data}};

\node (preprocessing) [process, below of=weather, yshift=-2cm] {\shortstack{Preprocessing}};
\node (weather_prep) [process, below of=preprocessing, xshift=-2.4cm] {\shortstack{Weather \\ Data \\ Prep.}};
\node (human_prep) [process, below of=preprocessing, xshift=0cm] {\shortstack{Human \\ Data \\ Prep.}};
\node (ground_prep) [process, below of=preprocessing, xshift=2.4cm] {\shortstack{Ground \\ Data \\ Prep.}};

\node (models) [model, below of=human_prep, yshift=-2cm] {\shortstack{Learning \\ Models}};
\node (lstm) [model, below of=models, xshift=-2.4cm] {\shortstack{LSTM \\ Networks}};
\node (cnn) [model, below of=models, xshift=0cm] {\shortstack{CNN \\ Models}};
\node (ground_model) [model, below of=models, xshift=2.4cm] {\shortstack{Ground \\ Data \\ Models}};

\node (ensemble) [ensemble, below of=cnn, yshift=-1.2cm] {\shortstack{Ensemble \\ Model}};
\node (transfer) [ensemble, below of=ensemble, yshift=-0.6cm] {\shortstack{Transfer \\ Learning}};

\node (parallel) [model, below of=transfer, yshift=-1.2cm] {\shortstack{Parallel \\ Modeling}};
\node (ffn) [model, below of=parallel, yshift=-0.6cm] {\shortstack{Feed \\ Forward \\ Networks}};

\node (output) [process, below of=ffn, yshift=-1.2cm] {\shortstack{Prediction \\ Output}};

\draw [arrow] (data) -- (satellite);
\draw [arrow] (data) -- (weather);
\draw [arrow] (data) -- (human);
\draw [arrow] (human) -- (ground);

\draw [arrow] (satellite) -- (weather_prep);
\draw [arrow] (weather) -- (weather_prep);
\draw [arrow] (human) -- (human_prep);
\draw [arrow] (ground) -- (ground_prep);

\draw [arrow] (preprocessing) -- (weather_prep);
\draw [arrow] (preprocessing) -- (human_prep);
\draw [arrow] (preprocessing) -- (ground_prep);

\draw [arrow] (weather_prep) -- (lstm);
\draw [arrow] (human_prep) -- (cnn);
\draw [arrow] (ground_prep) -- (ground_model);

\draw [arrow] (models) -- (lstm);
\draw [arrow] (models) -- (cnn);
\draw [arrow] (models) -- (ground_model);

\draw [arrow] (lstm) -- (ensemble);
\draw [arrow] (cnn) -- (ensemble);
\draw [arrow] (ground_model) -- (ensemble);

\draw [arrow] (ensemble) -- (transfer);
\draw [arrow] (transfer) -- (parallel);
\draw [arrow] (parallel) -- (ffn);

\draw [arrow] (ffn) -- (output);

\end{tikzpicture}
\caption{Schema of Proposed Models for Wildfire Prediction System}
\label{fig:wildfire-prediction-schema}
\end{figure}

\end{document}